\documentclass[journal,twoside,web]{ieeecolor}
\usepackage{generic}
\usepackage{cite}
\usepackage{amsmath,amssymb,amsfonts}
\usepackage{algorithmic}
\usepackage{graphicx}
\usepackage{algorithm,algorithmic}
\usepackage{hyperref}
\hypersetup{hidelinks}
\usepackage{textcomp}
\usepackage[table]{xcolor}
\usepackage{gensymb}
\usepackage{glossaries}
\usepackage[utf8]{inputenc}
\usepackage{pgfplots}
\usepackage{subcaption}
\usepackage{paralist}
\usepackage{changes}
\DeclareUnicodeCharacter{2212}{−}
\usepgfplotslibrary{groupplots,dateplot}
\pgfplotsset{compat=newest}
\def\BibTeX{{\rm B\kern-.05em{\sc i\kern-.025em b}\kern-.08em
    T\kern-.1667em\lower.7ex\hbox{E}\kern-.125emX}}
\newacronym{dof}{DoF}{degrees-of-freedom}
\newacronym{emg}{EMG}{electromyography}
\newacronym{AR}{AR}{Augmented Reality}
\newacronym{VR}{VR}{Virtual Reality}
\newacronym{BBT}{BBT}{Box-and-Blocks Test}
\newacronym{MPL}{MPL}{Modular Prosthetic Limb}
\newacronym{this}{ProACT}{Prosthetic Arm Control Testbed}
\newacronym{ROS}{ROS}{Robot Operating System}
\newacronym{IMU}{IMU}{Inertial Measurement Units}
\newacronym{hmd}{HMD}{Head-Mounted Display}
\newcommand*{\refname}{References}
\usepackage{soul}
\newcommand{\rev}[1]{\color{black}#1\color{black}}
\newcommand{\del}[1]{}

\begin{document}
\title{ProACT: An Augmented Reality Testbed for Intelligent Prosthetic Arms}
\author{Shivani Guptasarma, \IEEEmembership{Student member, IEEE}, and Monroe D. Kennedy III, \IEEEmembership{Member, IEEE}
\thanks{We acknowledge the support of this work by the Stanford Knight Hennessy Scholars, the Wu Tsai Human Performance Alliance at Stanford University and the Joe and Clara Tsai Foundation. We thank Gabriela Bravo Illanes and Subhechchha Paul for their valuable insights.}
\thanks{Both authors are with the Department of Mechanical Engineering, Stanford University, Stanford, CA 94305 USA (\{shivanig, monroek\}@stanford.edu).}}

\maketitle

\begin{abstract}
Upper-limb amputees face tremendous difficulty in operating dexterous powered prostheses. Previous work has shown that aspects of prosthetic hand, wrist, or elbow control can be improved through ``intelligent" control, by combining movement-based or gaze-based intent estimation with low-level robotic autonomy. However, no such solutions exist for whole-arm control. Moreover, hardware platforms for advanced prosthetic control are expensive, and existing simulation platforms are not well-designed for integration with robotics software frameworks. We present the Prosthetic Arm Control Testbed (ProACT), a platform for evaluating intelligent control methods for prosthetic arms in an immersive (Augmented Reality) simulation setting. \rev{We demonstrate the use of ProACT through preliminary studies, with non-amputee participants performing an adapted Box-and-Blocks task with and without intent estimation. We further discuss how our observations may inform the design of prosthesis control methods, as well as the design of future studies using the platform.} \del{Using ProACT with non-amputee participants, we compare performance in a Box-and-Blocks Task using a virtual myoelectric prosthetic arm, with and without intent estimation. Our results show that methods using intent estimation improve both user satisfaction and the degree of success in the task. } To the best of our knowledge, this constitutes the first study of semi-autonomous control for complex whole-arm prostheses, the first study including sequential task modeling in the context of wearable prosthetic arms, and the first testbed of its kind. Towards the goal of supporting future research in intelligent prosthetics, the system is built upon on existing open-source frameworks for robotics, and is available at \rev{\href{https://arm.stanford.edu/proact}{https://arm.stanford.edu/proact}}.
\end{abstract}

\begin{IEEEkeywords}
Assistive Robotics, Prosthesis, Augmented Reality, Human Enhancement.
\end{IEEEkeywords}
\section{Introduction}
\label{sec:introduction}
Prosthetic arms are typically controlled via signals acquired through \gls{emg} from the residual limb~\cite{mcleanEarlyHistoryMyoelectric2004,geethanjaliMyoelectricControlProsthetic2016}. As the number of input signals is inherently limited, these devices either have limited mechanical functionality, or have more \gls{dof} than can be simultaneously \rev{placed under direct independent control}\del{controlled by their users}. Dissatisfaction with functionality is a major reason for rejection and abandonment~\cite{yamamotoCrosssectionalInternationalMulticenter2019}, often within a month of use~\cite{salmingerCurrentRatesProsthetic2022}. 
Rejection rates vary between~$30-80\%$, and are the highest for limb loss near the shoulder~\cite{nationalacademiesofsciencesUpperExtremityProstheses2017, biddissUpperLimbProsthesis2007}. Many individuals with high-level and/or bilateral limb loss wish to use prosthetic arms, but find that, in their current state, these arms introduce more challenges than they alleviate~\cite{biddissUpperLimbProstheticsCritical2007, nationalacademiesofsciencesUpperExtremityProstheses2017}.
\subsection{Intelligent prosthetics}
\label{sec:ips}
In order to provide intuitive high-\gls{dof} operation with the low-\gls{dof} \gls{emg} input available, other sources of information \rev{may }\del{must} be utilized to automate low-level functions~\cite{swainAdaptiveControlSystem1980}. In the present work, as in the literature, we use the term ``intelligence" as a shorthand for such context-informed shared autonomy, as opposed to direct joint-level control. \\
The first intelligent control systems for transradial prostheses used computer vision and gaze tracking to identify the pose~\cite{dosen2011} and geometry~\cite{markovic_stereovision_2014, markovicSensorFusionComputer2015} of the target object, automating grasp preparation. \gls{AR} interfaces made the functioning of the device transparent to the user, who retained agency~\cite{guptasarmaConsiderationsControlDesign2023}. Later works used Inertial Measurement Units (IMUs) to track the residual limb~\cite{bennett2018, mouchouxArtificialPerceptionSemiautonomous2021a, starke2022} or the contralateral limb~\cite{volkmarImprovingBimanualInteraction2019}, and deep learning to predict joint angles~\cite{karrenbachImprovingAutomaticControl2022a}. In transhumeral prosthetics, intended elbow motion during reaching tasks has been estimated from the location of the reaching goal~\cite{martinNovelApproachProsthetic2014, garcia-rosasTaskspaceSynergiesReaching2020, segasMovementBasedProsthesisControl2024}, or using correlations between shoulder and elbow joints~\cite{kaliki2013, meradCanWeAchieve2018, ahmedTranshumeralArmReaching2023, ahmedSynergySpaceRecurrentNeural2023}. These approaches elegantly leverage the movement of the residual limb. \\
\begin{table*}[ht]
\caption{Simulation platforms for prosthetic arms. In principle, all platforms are customizable for various levels of limb loss, however, only ArmSym mentions shoulder disarticulation, using a touchpad as proxy input for a 2-\gls{dof} device.}
\setlength{\tabcolsep}{5pt}
\begin{tabular}{|c|c|c|c|c|c|c|}
\hline
Name & Purpose & Virtual arm type & Control method & Display & Environment & Access \\
\hline
JHU/APL VIE~\cite{armigerRealTimeVirtualIntegration2011} & Research & \acrshort{MPL} & \gls{emg} & 3D goggles & ODE, MATLAB & not available \\
JHU/APL miniVIE~\cite{RarmigerMinivieBitbucket} & User training & \acrshort{MPL} & \gls{emg} & Screen & MATLAB & open \\
MPI ArmSym~\cite{bustamanteArmSymVirtualHuman2021a} & Research & Barrett WAM & HTC Vive & Immersive \acrshort{VR} & Unity/C\# & open\\
HoloPHAM~\cite{baskarHoloPHAMAugmentedReality2017} & \rev{User t}raining & \acrshort{MPL} & \gls{emg} & Immersive \gls{AR} & Unity & not available \\
Virtual \acrshort{BBT}~\cite{hashimComparisonConventionalVirtual2021a} & \rev{User t}raining & Human avatar & \gls{emg} & Immersive \acrshort{VR} & Unity & not available \\  
ProACT (this work) & Research & \acrshort{MPL} & \gls{emg}, gaze, motion planning & Immersive \gls{AR} & ROS base, Unity interface & open \\
\hline
\end{tabular}
\label{tb:platforms}
\end{table*}
\rev{However, it is in }\del{Unfortunately, the problem of }\emph{whole-arm control}, as in the case of shoulder disarticulation or forequarter amputation, \del{receives little attention in the engineering literature}\rev{that users are most often dissatisfied with conventional myoelectric control methods~\cite{nationalacademiesofsciencesUpperExtremityProstheses2017, biddissUpperLimbProsthesis2007}. Unfortunately, this problem receives little attention in the engineering literature}. Krausz \emph{et al.} (2020) characterized the coordination between gaze and~$14$-channel \gls{emg} data from non-amputees, showing that gaze could be used with even a subset of \gls{emg} channels to predict grasping motions. Using shoulder adduction/abduction and biting motions as input, Togo \emph{et al.} (2022) demonstrated intelligent control of a custom-built 4-\gls{dof} whole-arm prosthesis~\cite{togoSemiAutomatedControlSystem2021} using gaze tracking to identify the target, and computer vision to plan reach and grasp motions. However, neither of these works was followed by studies extending the proposed methods to the control of higher-\gls{dof} wearable arms. \\
\rev{As the needs, capabilities, and preferences of myoelectric prosthesis users vary widely~\cite{brown2023}, we hypothesize that \emph{studying intelligent control methods for whole arms, where direct control inputs are the most limited, has the potential to give rise to insights and techniques benefiting people with all levels of amputation} in the future\footnote{along similar lines to the widespread adoption of predictive text, originally developed for use with disabilities~\cite{garay-vitoriaTextPredictionSystems2006}. As with text prediction, the expected variation in user preferences for intelligent prostheses (even within each level of amputation) underscores the need for large and detailed user studies.}.} \del{\emph{Research on intelligent control methods for whole arms can benefit people with all levels of amputation.} }
\subsection{Enhancing EMG inputs} 
Targeted Muscle Reinnervation (TMR)~\cite{kuikenTargetedReinnervationEnhanced2007} surgically increases the number and quality of \gls{emg} signals available from the residual limb, bringing myoelectric control within the realm of possibility for whole-arm prosthesis control. \del{However, w}\rev{W}hen successful, it creates \rev{upto}\del{at most}~$2$-$4$ independent channels of simultaneous control inputs~\cite{cheesboroughTargetedMuscleReinnervation2015}, \rev{which can be used with advanced pattern recognition techniques to perform coordinated joint motions~\cite{cheesboroughTargetedMuscleReinnervation2015}. }\del{; for comparison, the most advanced anthropomorphic prosthesis, the Modular Prosthetic Limb (MPL)~\cite{johannesChapter21Modular2020a} has~$26$-\gls{dof} with~$17$ actuators.} The feasibility \rev{and outcome } of TMR \del{also }depend\del{s} upon the conditions of limb loss, time elapsed, and surgeon training~\cite{felderFailedTargetedMuscle2022}. \rev{Meanwhile, the most advanced anthropomorphic prosthesis, the Modular Prosthetic Limb (MPL)~\cite{johannesChapter21Modular2020a} has~$26$-\gls{dof} with~$17$ actuators. } Thus, TMR alone \rev{(in spite of being the most prominent development thus far in whole-arm control) } cannot make whole-arm prosthetics intuitive to use \rev{in all cases.} \del{, yet it is the most prominent development thus far in whole-arm control. }\del{\emph{Intelligent control can benefit whole-arm prosthetics, both when TMR is possible and when it is not}.} 
\subsection{Immersive AR platforms}
Virtual environments are popular in training protocols for prosthesis users~\cite{perryVirtualIntegrationEnvironment2018a}, as well as in testing new methods of control for both prosthetics~\cite{simonPatientTrainingFunctional2012} and robot teleoperation~\cite{hertzHeadtoHeadComparisonThree2018, adamiEffectivenessVRbasedTraining2021}. Without procuring or fabricating hardware, researchers may study a wide range of devices and conditions in simulation using a single versatile setup, quickly and economically.\\
Immersive platforms provide more realistic results pertaining to human interaction, as compared to visualizations on flat screens~\cite{hepperleSimilaritiesDifferencesImmersive2023}. \gls{VR} systems such as ArmSym~\cite{bustamanteArmSymVirtualHuman2021a} and the Virtual Integration Environment or VIE~\cite{armigerRealTimeVirtualIntegration2011} take advantage of this. The HoloPHAM platform~\cite{baskarHoloPHAMAugmentedReality2017} uses \gls{AR} for direct \gls{emg} control training. 
Unlike~\gls{VR}, as mentioned in Section~\ref{sec:ips}, \gls{AR} is useful even as a component of real prosthetic systems, relaying information sensed from the environment (such as object recognition), information sensed from the user (such as gaze \rev{and contralateral hand motion}), controller decisions, and menu options. \rev{AR devices are also less likely to induce simulator sickness during long studies~\cite{vovk2018, sharmaAugmentedRealityProsthesis2019}, and in ambulatory tasks, give participants the confidence to move around the room naturally. Given the extensive software infrastructure that exists for the control of autonomous robotic manipulators, integrating it with the benefits of AR into a single platform would enable behavioral studies with intelligent prostheses across a broad scope. } \del{\emph{An \gls{AR} evaluation platform for intelligent prostheses would integrate such features with the extensive software infrastructure that exists for robot control}.} To the best of our knowledge (summarized in Table~\ref{tb:platforms}), such a platform does not exist. 
\subsection{Modeling of sequential tasks}
While object-specific properties (such as those mentioned in Section~\ref{sec:ips}) might be sufficient for transradial prosthetics, where the reaching motion is performed by the residual limb, whole-arm control is more ambiguous. It is common in robotics to use the semantics of a sequential task to predict likely interactions. There exist models predicting interactions in realistic settings~\cite{rodinPredictingFutureFirst2021}, as well as human-in-the-loop teleoperation experiments with simplified tasks that have explicitly defined rules~\cite{fuchsGazeBasedIntentionEstimation2021}. \emph{Modeling sequential tasks for action prediction is of great interest}, particularly for the control of whole-arm prosthetics; however, it does not seem to have been investigated in wearable prosthetic arms thus far.
\subsection{Problem Statement}
\label{sec:ps}
\begin{figure*}[tbhp!]
\centering
\includegraphics[width=\textwidth]{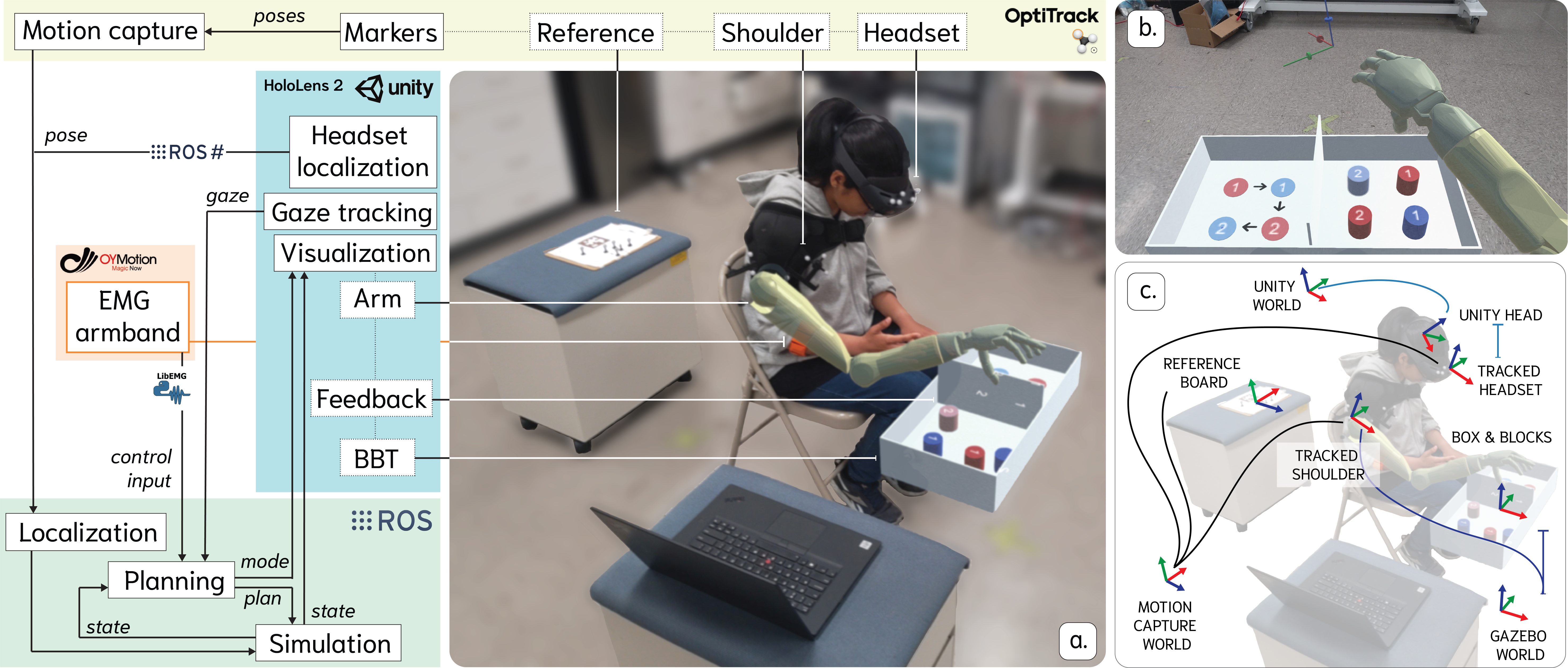}
\caption{(a) Components of the testbed, showing a non-amputee participant (pictured with permission) and virtual components (the arm, \gls{BBT}, and selection marker) as they would be seen by the participant. 
(b) The first-person view of the \gls{BBT}. (c) The mixed-reality scene is localized with respect to the room, as explained in detail in Section~\ref{sec:sysarch}.}
\label{fig:sysarch}
\end{figure*}
The broader research question to be investigated is whether users of powered prosthetic arms\del{-- with all levels of amputation --} can perform manipulation tasks better and more comfortably, if direct control is augmented with the use of passively revealed cues from the user, sensing and interpretation of the scene and task context, autonomous control of some aspects of joint motion, and \gls{AR} feedback. In order to address this question, an evaluation platform is needed which:
(\textbf{1}) is cost-efficient, time-efficient, and versatile, accommodating studies of partial as well as complete limb loss, with amputee as well as non-amputee participants,
(\textbf{2}) is reasonably physically accurate,
(\textbf{3}) is \rev{informative} for behavioral studies (limiting distortion of participant behavior),
(\textbf{4}) is accessible to robotics researchers (compatible with existing task and motion planning frameworks, computer vision libraries, intention prediction models and other algorithms developed for robotics), and
(\textbf{5}) is compatible with a variety of input modalities (\gls{emg}, joysticks, cameras, gaze trackers) and feedback modalities (tactile actuators, speakers, and a visual display).
\subsection{Contributions}
We present the \gls{this}, integrating motion planning and control in \gls{ROS}~\cite{ros} with \gls{emg} control and an immersive \gls{AR} interface (Fig.~(\ref{fig:sysarch})), in order to meet the requirements described in Section~\ref{sec:ps}. The contributions of this work are:
\begin{enumerate}
    \item an open-source platform for the robotics and prosthetics research communities to study intelligent prostheses in immersive simulation,
    \item \rev{two exploratory user studies of intelligent control for a high-\gls{dof} wearable prosthetic arm, demonstrating the usability of the above platform for non-amputees, and}
    \item \rev{preliminary data suggesting possible advantages of intelligent control design for whole-arm prostheses, and highlighting concerns relevant to the design of such interfaces.}
    \del{\item a user study of intelligent control for a high-\gls{dof} wearable prosthetic arm, made possible by the above platform, and
    \item the inclusion of sequential modeling of a task, as an exploratory component of the study.}
\end{enumerate}
Details of the design and implementation of the testbed and the user stud\rev{ies }\del{y} are provided in Section~\ref{sec:methods}\del{ below}. Results \del{from the study }are presented in Section~\ref{sec:results} and discussed in Section~\ref{sec:discussions}, together with limitations and possible modifications. Section~\ref{sec:conclusion} summarizes \rev{outcomes }\del{ conclusions} from the present work and \del{possible }future \rev{directions}\del{ applications}.
\section{Methods}
\label{sec:methods}
The details of the platform are given in Section~\ref{sec:sysarch}. Section~\ref{sec:task} explains the design of the evaluation task in light of existing tests. The control methods used in the study are described in Section~\ref{sec:interface}. Finally, Section~\ref{sec:study} reports the design of the user stud\rev{ies}\del{y}.
\subsection{System architecture}
\label{sec:sysarch}
The platform (Fig.~(\ref{fig:sysarch}a)) is based on \gls{ROS} Noetic, running on an Ubuntu 20.04 \rev{computer } (Intel Core i9-12900K CPU,~$64$~GB RAM). A real-time dynamics simulation ($1$~kHz update rate,~$1$~ms time step) in Gazebo\footnote{Unity was not used for dynamics, since it prioritizes visual realism and stability over physical accuracy (By default, Unity and Gazebo rely on the physics engines PhysX and ODE or Open Dynamics Engine respectively~\cite{erezSimulationToolsModelbased2015}). } is displayed to participants (Fig.~(\ref{fig:sysarch}b)) at~$60$~Hz, via a mixed-reality application written in Unity~2020.3 and deployed to a Microsoft HoloLens~2 \gls{hmd}. \rev{For localizing the arm in space, a Windows 10 computer (Intel Core i9-9900KS CPU,~$64$~GB RAM) transmits motion capture data from an~$8$-camera OptiTrack system over a 5G WiFi network. The open-source ROS\# libraries~\cite{bischoffm_2019_06} are used to communicate between ROS and Unity. }\\
\textbf{\del{Communication}\rev{Control}}: \del{The open-source ROS\# libraries 
are used to communicate between ROS and Unity.} An OyMotion \gls{emg} armband worn on the forearm sends control inputs over a Bluetooth Low Energy (BLE) connection. This allows for quick setup (no individual electrode placement), while approximating the few sequential control inputs \rev{that might be } available to a shoulder disarticulation amputee. We use \rev{the }\del{recently-released }LibEMG libraries~\cite{libemg} to classify~$5$ discrete hand gestures via \rev{Singular-Value Decomposition-based } Linear Discriminant Analysis (LDA)\rev{, publishing classes as ROS messages\footnote{A message is published only if the classifier returns the same output three times consecutively.}, subscribed to with a queue length of~$1$ (to use the latest input). A Mean absolute value, zero crossings, slope sign change and waveform length are extracted \del{as features }from data acquired at~$1000$~Hz, using a sliding window size of~$100$ samples with a~$50$-sample increment between windows~\cite{GitHubLibEMGLibEMG_Snake_Showcase}.} \\ 
\textbf{Localization}: The base of the virtual prosthetic arm tracks motion capture markers attached to a shoulder brace (Fig.~(\ref{fig:sysarch}c))\rev{, with the ROS tf2 transform library being used to manage coordinate frame transformations over time}. At the start of each trial, the transform between the Unity and motion capture reference (world) frames is found using the \gls{hmd}. \rev{The \gls{hmd} } \del{which} is tracked in both systems (via built-in localization and attached markers, respectively). The box and blocks are displayed at a fixed height measured from the shoulder in an upright posture, at a fixed horizontal displacement from the reference board. Having thus located the Gazebo world frame in the physical space, all virtual objects are displayed in their correct poses relative to the Unity world \rev{(resetting takes under a minute)}.  \\ 
\textbf{Device}: The default arm used is the \gls{MPL}, a research platform for sophisticated anthropomorphic whole-arm prostheses. The modern version of the arm, unlike the online model~\cite{Gazebo_modelsMpl_right_armMaster}, includes shoulder flexion~\cite{johannesChapter21Modular2020a}, hence we add a \gls{dof} at the base. The online~$26$-\gls{dof} model, unlike the~$26$-\gls{dof} model described in the chapter~\cite{johannesChapter21Modular2020a}, has a~$20$-\gls{dof} hand (rather than~$19$). The addition of the shoulder flexion joint results in a~$7$-\gls{dof} arm and~$20$-\gls{dof} hand. Such a high-\gls{dof} prothesis, impossible to control completely with~$2$-$4$ sequential inputs, is more suitable for studying the impact of intent estimation than one which may have a one-to-one mapping of input channels, but may not have sufficiently high~\gls{dof} to perform complex tasks. 
\subsection{Task design}
\label{sec:task}
The \acrlong{BBT}~\cite{cromwell1960occupational} is a well-established measure of upper-limb dexterity for stroke patients in recovery. It has been modified for specific conditions such as Parkinson's disease~\cite{hwangRelationshipManualDexterity2016}, cerebral palsy~\cite{liangMeasurementPropertiesBox2021a, zapata-figueroaAssessmentManualAbilities2022}, fibromyalgia~\cite{cannyReliabilityBoxBlock2009}, and prosthesis use~\cite{kuikenTargetedReinnervationEnhanced2007a, hebertCaseReportModified2012a, kontson_targeted_2017}. The task requires participants to transport blocks from one side of a box to another, over a partition. The partition and walls of the box ensure that some extent of complex manipulation is demanded. Virtual versions of the test have been developed for many of these conditions\del{as well}~\cite{everardConcurrentValidityImmersive2022a, onaEvaluatingVRbasedBox2019a, hashimComparisonConventionalVirtual2021a}. In the original test,~$150$ blocks are heaped on one side and the score is defined by the number of blocks moved in a fixed time~\cite{mathiowetzAdultNormsBox1985a}. Modifications \rev{to evalute }\del{for the evaluation of }prosthesis use in particular, the mBBT~\cite{hebertCaseReportModified2012a} and tBBT~\cite{kontson_targeted_2017}, decrease the number of blocks, and specify the initial positions, place locations, and sequence of interaction. The score\del{, in these tests,} is defined by the time taken to move all the blocks. The controlled sequence reduces variability between subjects introduced by choice, and controlled place locations make it possible to study the entire task including precis\rev{e actions}\del{pick and place}.\\
\textbf{Modifications to task}: We make the following further modifications:
\begin{inparaenum}[(1)]
    \item The size of blocks is increased to mitigate visual occlusion by the hand.
    \item The number of blocks is further decreased and spacing increased, so the box dimensions remain reachable, with the walls continuing to constrain reaching.
    \item\del{ As real-time, realistic contact simulation is difficult, c}\rev{C}ylindrical blocks are used, where grasp shape is agnostic to approach direction.
    \item Numbered blocks of two different colors are used. Color and number are used as a proxy for the semantics of a sequential task, with only the color being taken into account as described in Section~\ref{sec:interface}. Similarly to the mBBT and tBBT, \rev{the order is specified}\del{participants are instructed to move the blocks in a specific order}, but the initial configurations are counterbalanced over \rev{their }possible arrangements\rev{, as }\del{. This is necessary because }the difficulty of reaching different regions of the box varies widely and it is infeasible to get data for all blocks in all trials (as explained below).\\
\end{inparaenum}
\textbf{Modifications to format}: Various versions of the test require participants to sit~\cite{mathiowetzAdultNormsBox1985a} or stand~\cite{kontson_targeted_2017}, and score performance based on either the number of blocks moved in a fixed period of time~\cite{mathiowetzAdultNormsBox1985a} or the time taken to move all the blocks~\cite{kontson_targeted_2017}. For the present context, these parameters were determined based on the following observations from pilot studies:
First, when standing and moving freely, \del{pilot }participants were able to achieve nearly identical performance in all methods by using whole-body motion for reaching. Seating \rev{them }\del{participants} forces them to operate the \gls{MPL} arm, making comparisons possible. 
Second, mistakes reflect the quality of control, but affect subsequent actions. Multiple trials are needed so that incidental mistakes do not \rev{impact }\del{result in the loss of }all data for some participant-method combinations.
Finally, the onset of fatigue has a strong effect on performance\del{ and time taken}. Extended \gls{hmd} use also leads to slight headaches, eye strain, heating of the device, and battery depletion. The Hololens 2 battery lasts for~\rev{$2$-$3$ }\del{two} hours of continuous operation. Participants' skill in the task varies widely, some spending~$5$-$10$ minutes \emph{per block} with the more difficult methods. Accounting for the demonstration and training phase, and the need for multiple trials\del{ with all methods} within a single session of reasonable duration, \del{the duration of }each trial is limited to~$5$ minutes.\\
\textbf{Anthropometric adjustment}: The MPL is a large device. Scaling to participant height would require editing of not only geometric but inertial parameters, which is not realistic (real prosthetic arms have weight and dimensions determined by functionality, design and industrial standardization). Hence, rather than scale the arm or the \gls{BBT}, the table height is set at~\rev{$0.43$}~m below the \rev{measured location of the \gls{MPL} } shoulder before starting each trial. This distance was determined during pilots, such that it would be possible to comfortably reach all the blocks. \rev{The partition is aligned with the participant's sagittal plane using the reference board, which can be slightly corrected based on the participant's reporting if necessary. The horizontal distance between the participant and the box was adjusted by participants in Study 1 (to be as close as possible without intersecting their legs), and was fixed at~$0.6$~m to the box center in Study 2 (for better consistency across participants, considering the fixed size of the arm).}
\subsection{Feedback and control}
\label{sec:interface}
\begin{figure}[thpb]
\centering
\includegraphics[width=0.5\textwidth]{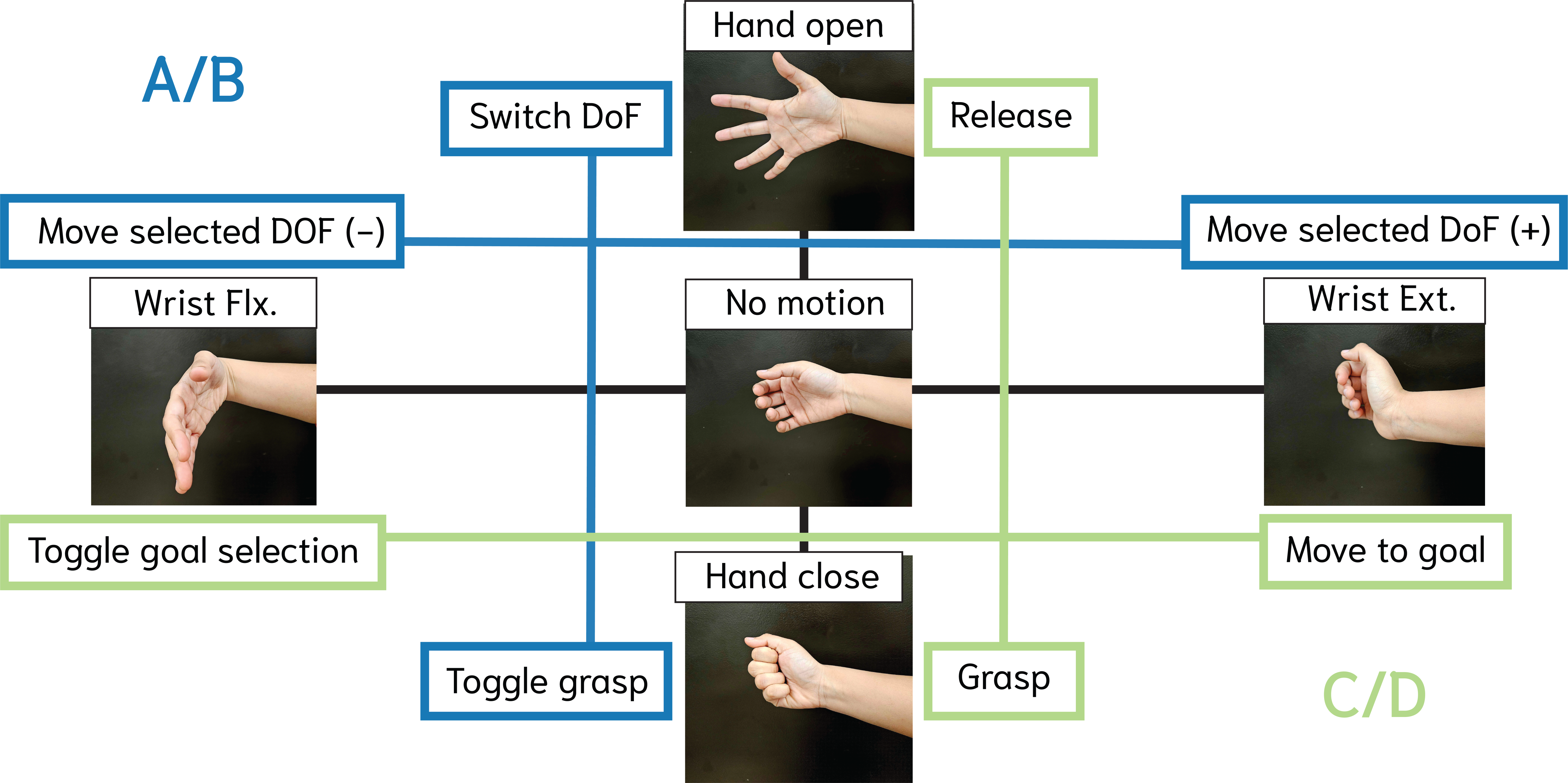}
\caption{The four control methods (A-D) mapping \gls{emg} inputs to \gls{MPL} motion.}
\label{fig:ctrl}
\end{figure}
Two hand poses are pre-programmed as ``open" and ``closed", the closed pose being a precision circular grasp approaching the block diameter. \rev{In place of simulating contact physics, t}\del{T}he Gazebo grasp plugin~\cite{GitHubJenniferBuehlerGazebopkgs} is used to rigidly attach objects to the fingertips based on proximity. Participants are made aware of this implementation. Distinct sounds are played through the \gls{hmd} when contact is made or lost. The~$5$ hand gestures 
are mapped to the movements of the~$7$ joints of the \gls{MPL} arm, and the opening and closing of the \gls{MPL} hand. These movements are referred to below as HO (hand open), HC (hand close), WF (wrist flex), WE (wrist extend) and NM (no motion). Four methods of control are compared (Fig.~\ref{fig:ctrl}), spanning the spectrum between direct and assisted control. \\
\textbf{Direct joint control (A)}: 
Controlling one joint at a time, the participant uses HO to cycle through the joints in the sequence: shoulder flexion/extension, shoulder adduction/abduction, shoulder internal/external rotation, elbow flexion/extension, wrist pronation/supination, wrist ulnar/radial deviation, and wrist flexion/extension. The name of the selected joint is displayed as a virtual text mesh. WF/WE move the selected joint in +/- directions, at pre-set speeds, by sending incremented joint goals to MoveIt joint trajectory controllers. HC toggles the state of the hand.\\
\textbf{Direct end-effector control (B)}: Computing inverse kinematics with the Kinematics and Dynamics Library (KDL)~\cite{kdl-url}, joint goals are assigned such that the end-effector (the most distal wrist link) moves along the selected direction in the~$6$-\gls{dof} task space, either translating along or rotating about one of the axes of a coordinate frame attached to the base of the \gls{MPL} (the ``tracked shoulder" frame in Fig.~(\ref{fig:sysarch}c), which is also displayed in front of the participant as in Fig.~(\ref{fig:sysarch}b), highlighting the selected direction). HO cycles through~$X, Y, Z$ translation and rotation \gls{dof}s. 
WF/WE move along the selected direction, and HC toggles the state of the hand. \\
In the general case, the Moore-Penrose inverse of the~$6\times 7$ Jacobian matrix is used to find the joint-space goal:
\begin{align}
    \boldsymbol{q}_{t+1}= \boldsymbol{q}_t + \boldsymbol{J}^\# \begin{bmatrix} \dot{\boldsymbol{x}} & \boldsymbol{\omega} \end{bmatrix} ^\top \Delta t.
\end{align}
The starting configuration is non-singular. An upper limit on the condition number of~$\boldsymbol{J}$ is implemented to limit elbow extension. Any other instability caused by proximity to singularities is allowed as being characteristic of this method.\\
\begin{figure}[thpb]
\centering
\includegraphics[width=0.45\textwidth]{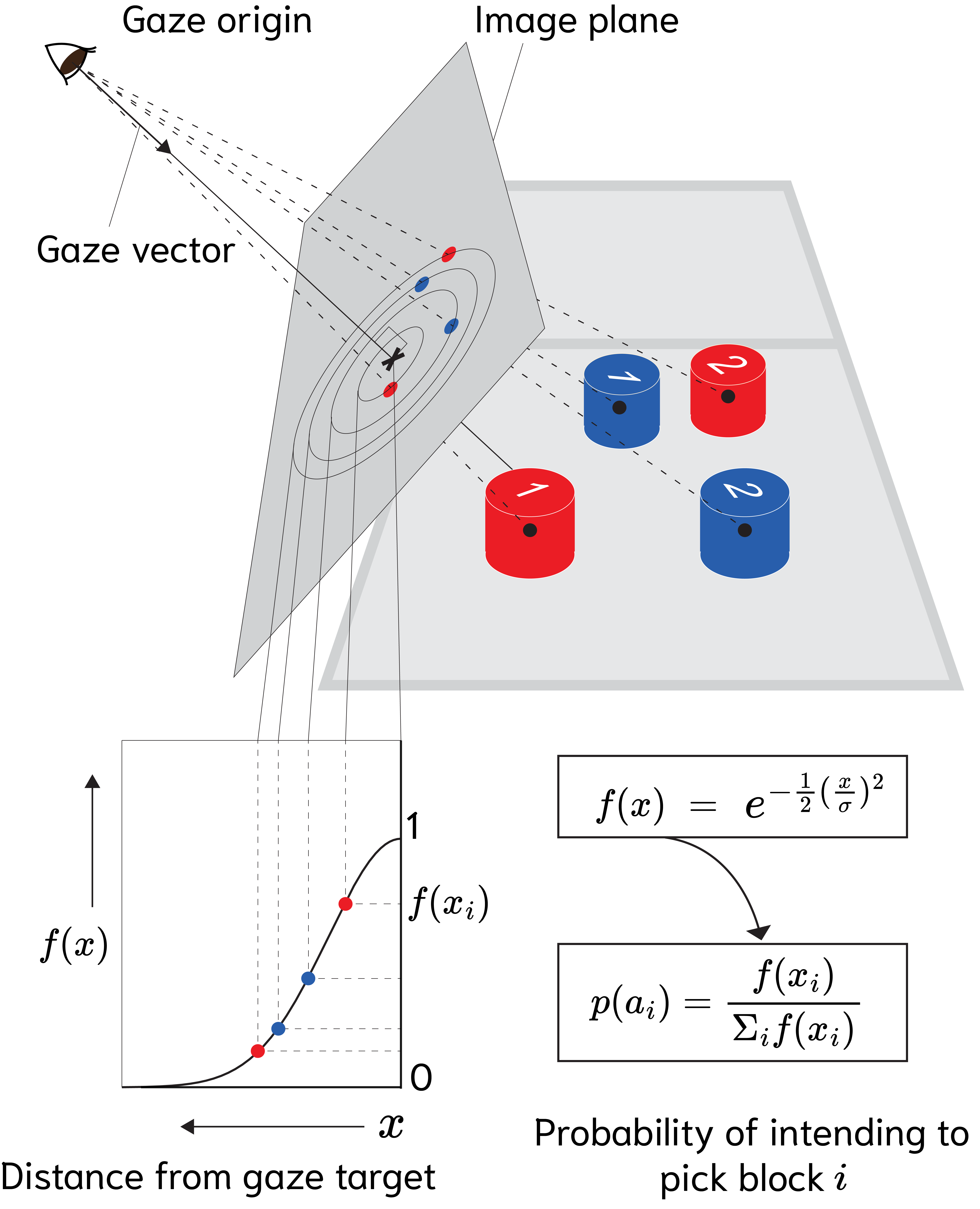}
\caption{The image plane is defined as a plane normal to the gaze vector. Each block is assigned a probability based on its distance from the gaze target in the image plane, thus: a bivariate Gaussian function centered at the gaze target is evaluated at the centers of all blocks, and the resultant values normalized by their sum. This simplistic calculation is sufficient for our aims in this study.}
\label{fig:gauss} 
\vspace{-1em}
\end{figure}
\textbf{Gaze-assisted control (C)}:
The participant's gaze (tracked by the \gls{hmd}\rev{, within a~$1.5\degree$ error}) is used to infer a target, as shown in Fig.~(\ref{fig:gauss}), which is highlighted in AR. Participants use WF to ``lock" the selection and WE to execute motion towards the selected object. For placing, a dark marker tracking gaze appears on the target side of the box. WF locks the position of the spot and WE executes motion towards it. Motion is based on trajectories generated by sample-based planners selected by the Open Motion Planning Library (OMPL), with asynchronous execution, so that NM immediately stops the motion. A new motion plan is attempted every time a change to WE is detected. If WE is held, the existing plan is used. Participants are made aware that, if the base of the arm is moved during execution, a new planning action (NM-WE) might be required. The selection marker changes color upon locking and unlocking the gaze-based selection. When the goal is unreachable, the planner sets an intermediate goal.\\ 
\textbf{Context-assisted control (D)}:
A modification of method C; in addition to gaze, highlighting is biased by the color of the object. 
The above is a simplistic proxy for knowledge of the semantics and state of a task influencing the inference of user intent, in addition to cues gathered from the user's own behavior. In general, let us consider the prediction of the probability of the action~$a_i$ being intended by the user, given cues~$g$ observed from the user (such as gaze) and the state of the environment or task~$s$ (in this case, the color of the block just dropped); we may perform Bayesian inference:
\begin{align}
    p(a_i | g, s) = \frac{p(g | a_i, s) p(a_i | s)}{p(g | s)},
\end{align}
making the following simplifying assumptions/observations:
\begin{inparaenum}[1)]
    \item The denominator,~\mbox{$p(g|s) = \Sigma_i p(g, a_i | s)$}, is the same for all~$i$ and will not affect the ranking of the actions (blocks to pick). This quantity can therefore be safely neglected.\\
    \item The state of the task~$s$ affects~$g$ only through its effect on~$a_i$, i.e., the probability distribution for the target of the gaze vector in the image plane is completely determined by the identity and location of the block to be picked, regardless of the state of the task. Therefore, if there are~$m$ states,~\mbox{$p(g | a_i, s) = \frac{1}{m}~p(g|a_i) $}. We assume that the probability distribution of gaze in the image plane assumes a normal form centered about the intended target block, hence,~\mbox{$p(g | a_i)\propto e^{- \frac{x_i^2}{2 \sigma^2}} $}, where~$x_i$ is the distance in the plane between the center of the block in question and the gaze target.\\
    \item We assume the distribution of intention conditioned on state~$p(a_i|s)$ as shown in Table~\ref{tb:dprobs}~\footnote{If the previous block was dropped prematurely and is being reattempted, or the participant is making a mistake, the system still allows them to select blocks of the same color by gazing directly at them.}. \\
\end{inparaenum}
\begin{table}
\caption{Probability weighting~$p(a_i|s)$ for each block in Method~D, conditioned on previous block color. The columns, left to right, are in the sequence specified for the task.}
\label{tb:dprobs}
\begin{tabular}{|p{70pt}|p{30pt}|p{30pt}|p{30pt}|p{30pt}|}
\hline
Previous block color & Red \#1 & Blue \#1 & Red \#2 & Blue \#2 \\
\hline
Red & 0.125 & 0.375 & 0.125 & 0.375 \\
Blue & 0.375 & 0.125 & 0.375 & 0.125 \\
\hline
\end{tabular}
\end{table}
Thus assigning a ranking to the blocks based on both gaze and state, the block ranked as having the highest probability of being the intended target at any given time is highlighted.
\subsection{Study design}
\label{sec:study}
\rev{The modified \gls{BBT} described in Section~\ref{sec:task} was used as a basis for two user studies, to test the reliability of the platform and gain preliminary insights into the methods. Participants } gave informed consent under the oversight of an Institutional Review Board\footnote{Protocol~65022, approved by the Administrative Panels for the Protection of Human Subjects at Stanford University, on February 29, 2024} and were all right-handed non-amputees.\\
\rev{In Study~1,~$8$ participants (ages~$22$-$30$;~$5$M/$3$F) used all~$4$ methods for~$3$ trials each, within a single~$120$-minute session. } The \rev{method } sequence was counterbalanced \rev{ using a partial Latin square (Table~\ref{tb:seq}), to evenly distribute the effects of skill transfer and fatigue. D}\del{ As shown in (Table~\ref{tb:seq}), alternating d}irect (A, B) and assisted (C, D) methods \rev{were presented alternately}, so that the assisted methods, with similar participant controls, were never used contiguously. \\
\rev{In Study~2, 
methods~A and~D were examined at greater length, with~$6$ male participants (ages~$22$-$31$\footnote{P1 from Study~1 volunteered as P2 in Study~2 and was given all the same instructions as other participants. Other participants were na\"ive.}) performing~$12$ trials with a single method in each of two~$120$-minute sessions~($1$-$4$ days apart). The goal was to collect richer data per method, as high failure rates had impacted quantitative data collection in the shorter Study~1, and to alleviate fatigue. In order to get sufficient data within a limited duration, all participants were administered the methods in the sequence D-A, while recognizing that a skill transfer effect might be seen on the second day. This decision was based on the relatively greater difficulty seen with method~A in Study~1, when participants were not given enough time to acclimate to the task.} \\
\rev{\emph{Together, the two studies show how ProACT can be used to (1) quickly compare and prototype multiple interfaces for intelligent prostheses, and (2) perform detailed investigations on parameters relating to movement and task completion.}}\\
\del{Eight right-handed non-amputees (ages~$22$-$30$;~$5$M, $3$F) gave informed consent for a within-participants study 
under the oversight of an Institutional Review Board\footnote{Protocol~65022, approved by the Administrative Panels for the Protection of Human Subjects at Stanford University, on February 29, 2024.}. Each participant attempted~$3$ trials of the modified \gls{BBT} described in Section~\ref{sec:task} with each of the~$4$ methods. The sequence was counterbalanced\rev{ using a partial Latin square, to evenly distribute the effects of skill transfer and fatigue. As shown in} (Table~\ref{tb:seq}), \del{alternating }direct (A, B) and assisted (C, D) methods \rev{were presented alternately}, so that the assisted methods, with similar participant controls, were never used contiguously. \\}
\begin{table}
\caption{Counterbalanced sequence of methods \rev{in Study~1}}
\label{tb:seq}
\begin{tabular}{|>{\centering\arraybackslash}p{0.1\textwidth}|>{\centering\arraybackslash}p{0.06\textwidth}|>{\centering\arraybackslash}p{0.06\textwidth}|>{\centering\arraybackslash}p{0.06\textwidth}|>{\centering\arraybackslash}p{0.06\textwidth}|}
\hline
Participant & 1 & 2 & 3 & 4 \\
\hline
1 & A & ~C* & B & D \\
2 & C & B & ~D* & A \\
3 & A & D & B & C \\
4 & D & B & C & A \\
5 & B & D & A & C \\
6 & D & A & C & B \\
7 & B & C & A & D \\
8 & C & A & ~D* & ~B* \\
\hline
\multicolumn{5}{p{0.46\textwidth}}{The trials marked * diverged from this plan as follows, due to technical issues, deterioration of EMG classification or difficulty training in certain trials: participant 1 repeated C after a gap of several days; participant 2 repeated D similarly; participant 8 performed trials with C and D in a separate session on the next day after A and B.}
\end{tabular}
\vspace{-1em}
\end{table}
\textbf{Protocol}:
After filling a short survey regarding their prior experience with human-robot interaction devices, participants received a standardized verbal description of the task. \rev{They followed a screen-guided training procedure for EMG gesture classification, holding each gesture for~$3$~seconds and performing the sequence of gestures twice, while seated with horizontal unsupported forearm (as it would be held during the trials). Classification performance was evaluated using an on-screen cursor control test. During this familiarization period, participants also had a chance to adapt their gestures for clarity. In Study~1, the protocol was continued when participant and experimenter were both satisfied with classifier performance. In Study~2, participants were asked (at a minimum) to perform all gestures in a prescribed sequence, three times consecutively, without misclassification. } \del{They trained and tested the EMG classifier using \rev{an on-screen cursor control test}\del{a simple custom game} shown on a screen.} 
Next, participants wore the \gls{hmd}, \rev{performed gaze calibration, } and were seated on a chair\del{, centering it relative to the box and as close to it as possible without intersection. They then} \rev{for }\del{received} demonstrations\del{ for methods A, B and C}. \rev{In Study~1, f}or A and B, they were allowed to try moving and switching through \del{each}\rev{all } \gls{dof}, as well as toggling the hand state, until they understood the operation conceptually\footnote{Based on pilot studies which showed that participants found them tedious and confusing at first, performing a full pick and place with these methods would have \rev{unpredictably } extended the demonstration phase of the study\del{ to unpredictable lengths}.}. For C, since instructions were dependent on the stages of the task, they were guided verbally through one pick and place. \rev{If necessary, } \gls{emg} training was repeated\rev{ , and a~$5$-minute break given}\del{if necessary}, before beginning trials. \rev{In Study~2, due to the additional time available, they were allowed to complete two block transfers with each method (D/A) as practice, followed by a~$5$-minute break, and with another~$2$-minute break after~$6$ of~$12$ trials\footnote{In Study~1,  natural breaks from the task occurred during the intermediate surveys. Additional breaks were given if requested, but were not enforced.}. In Study~1, a}~$5$-minute countdown timer was shown on a laptop screen placed in front\rev{, which was omitted in Study~2 so as to obtain gaze tracking data pertinent to the task alone}. Participants \rev{in both studies } were asked to prioritize completion rate (maximize number of blocks transferred before timeout), speed (minimize total time taken), and accuracy (minimize distance between actual and desired placement location), in that order. \rev{Importantly, in Study~1, they were explicitly told that they were free to move their upper body, while in Study~2, they were asked to limit such motion, actively trying to use the provided methods except when necessitated by reachability constraints. } \del{\gls{emg} training was repeated as necessary, if it deteriorated during trials.} \rev{Trials were monitored through Gazebo and RViz rather than streaming the participant's view, so as not to impede communication and rendering on the HMD.}\\
\textbf{Surveys}: 
\del{The commonly-used NASA-TLX~\cite{hartDevelopmentNASATLXTask1988}} \rev{In Study~1, long questionnaires were } \del{was} avoided, \del{considering that its length and repetition would encourage} \rev{ to discourage ``}satisficing\rev{"}~\cite{krosnickSatisficingSurveysInitial1996} \rev{due to time pressure and fatigue}\del{in a tiring task}~\cite{choiCatalogBiasesQuestionnaires2004, hambySurveySatisficingInflates2016}. Instead, a custom~$5$-point,~$3$ question survey was given, querying the perceived \rev{levels of } ease, sense of agency and speed\del{ of performing the task}. At the end of the experiment, participants ranked the methods based on the same questions and were asked which method they would choose if they could ``\rev{do }\del{play} one last \del{game}\rev{trial } with any one method", \rev{with }\del{and} the reason for their choice. \rev{In Study~2, which was split into separate sessions and allowed for more time, the Pros-TLX instrument~\cite{parrToolMeasuringMental2023} (kindly brought to our attention by a reviewer) was used instead, and subjective comments noted.}
\section{Results}
\label{sec:results}
\rev{Selected results of interest from the two studies are presented below, with the complete data available on the project website.}
\subsection{Study 1}
\del{
Analyses were planned for:
\begin{inparaenum}[(1)]
    \item block outcomes (success rate in reaching targets),
    \item pick duration and place duration when successful, and
    \item placement accuracy (minimum distance achieved from target position over the course of the trial).
\end{inparaenum}
}
Block outcomes \rev{(success rate in reaching targets) }dominated the results, with C and D outperforming A and B uniformly across participants. \del{Due to the high failure rates, only sparse data were available from A and B for \del{the remaining}\rev{other } metrics. Therefore, quantitative results are presented for success rates, and the remaining results reported through visualizations.}
\begin{figure}[thpb]
\centering
\includegraphics[trim={4.5cm 0cm 5cm 2.2cm},clip,width=0.49\textwidth]{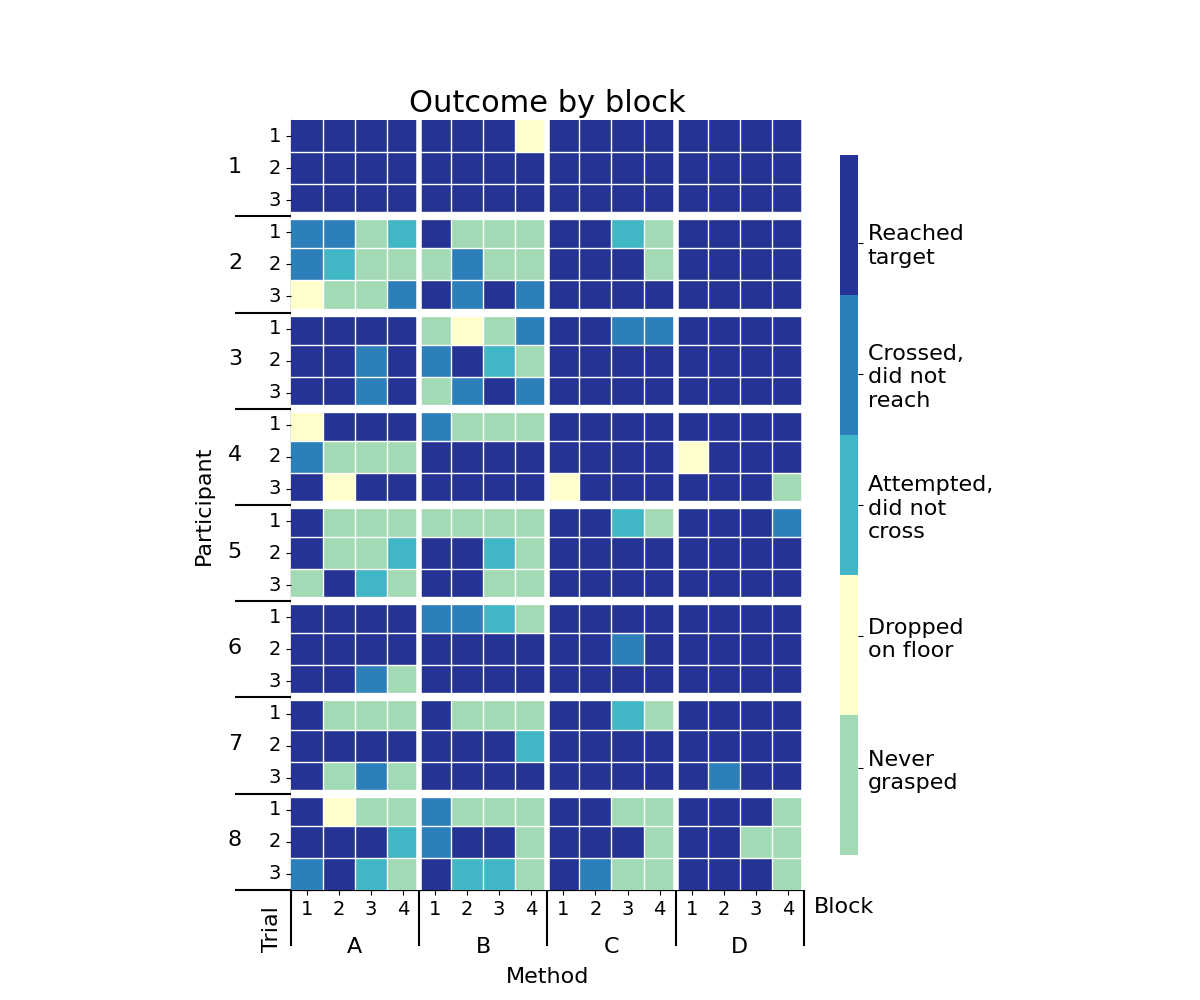}
\caption{The outcome shown for each individual block in the experiment \rev{(Study~1)}.}
\label{fig:fatedetail}
\end{figure}
\del{
\begin{figure}[hbpt]
\vspace{-1em}
\centering
    \includegraphics[width=0.45\textwidth]{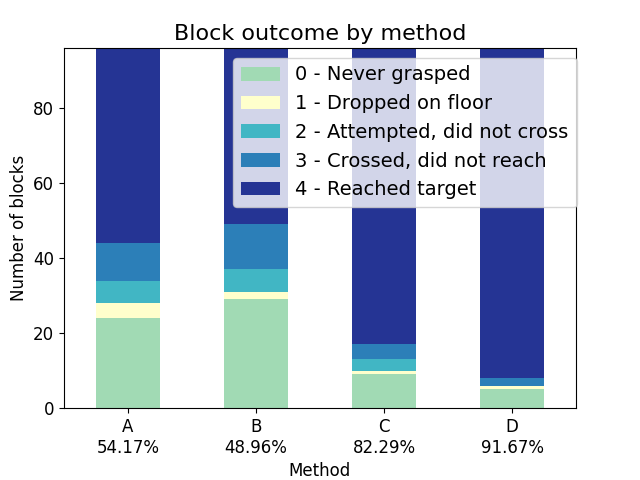}
    \caption{Pooling participant data \rev{from Study~1}, the number of blocks for each outcome, shown sorted by method.~$96$ blocks were presented with each method. The percent of blocks that successfully reached their targets is noted for each method.}
    \label{fig:fates}
\end{figure}
}
\subsubsection{Success}
Each participant performed~$12$ trials, and \rev{thus } had the opportunity to interact with~$48$ blocks in total. Each block among these had exactly one of the outcomes shown in Fig.~(\ref{fig:fatedetail}).
There was considerable variation in performance across participants. We can make the following observations:
First, \del{from Fig.~(\ref{fig:fatedetail}), }the two methods with the highest number of successful transfers (out of~$3 \times 4$) are C and D, for every individual participant. Second, most participants had difficulty using B at first ($5$ of them had no success at all in the first trial with B), but improved in subsequent trials. Third, in the aggregated data from all participants (Table~\ref{tb:success}), the mean and median number of successes per trial (over~$8 \times 3$ trials, with~$4$ being the maximum success rate per trial) are both higher with C and D. The range and standard deviation are both smaller. Thus, C and D were more successful on average, and more uniformly so, even across participants.
Finally, \del{Fig.~(\ref{fig:fates}) shows the distribution of outcomes within each method, pooling all the blocks moved by all participants. F}\rev{f}or every \rev{individual } failure outcome, fewer blocks had that outcome with C, D than with A, B. Method B had the highest number of blocks which crossed the partition without reaching the target. The most blocks were dropped on the floor with A. 
\begin{table}
\caption{Trial success rates aggregated over participants \rev{(Study~1)}.}
\label{tb:success}
\begin{tabular}{|p{25pt}|p{25pt}|p{25pt}|p{30pt}|p{30pt}|p{30pt}|}
\hline
Method & $\mu$ & $\sigma$ & Minimum & Maximum & Median \\
\hline
A & 2.21 & 1.50 &  0 & \textbf{4} & 2.5 \\
B & 1.96 & 1.60 &  0 & \textbf{4} & 2.0 \\
C & 3.29 & 0.95 &  1 & \textbf{4} & \textbf{4.0} \\
D & \textbf{3.67} & \textbf{0.56} &  \textbf{2} & \textbf{4} & \textbf{4.0} \\
\hline
\multicolumn{6}{p{0.95\columnwidth}}{The mean~$\mu$, standard deviation~$\sigma$, median and range values presented are for the number of blocks (out of the~$4$ available in each trial) that were successfully moved to the target region, aggregated over the~$8 \times 3$ trials performed with each method}
\end{tabular}
\vspace{-2em}
\end{table}
\subsubsection{Transfer timings and accuracy}
\label{sec:timings}
In every trial, participants had the freedom to operate the arm while moving its base (their shoulder), or to simply move the base without actively using the method. The pick and place actions were annotated with this information, based on the joint angle trajectories of the \gls{MPL}. Pick duration was defined as the time from the release of the previous block (or start of the trial) until contact was made. Figure.~(\ref{fig:pick}), shows the time taken \del{for}\rev{to perform }those pick\del{s}\rev{ing actions } which were stable enough for the block to eventually cross the partition~($266$ in number; although, for blocks grasped multiple times, the first time was used). The duration of a successful place action was defined as the time from grasping to release, under the conditions that (a) the block crossed the partition during this period, and (b) the target region (within~$0.05$~m\rev{, or one block diameter, } of the target) was entered before the block was released. These durations~($258$ in number) are shown in Fig.~(\ref{fig:place}). \del{Finally, for those blocks which successfully reached the target regions~($266$ in number, including those dropped slightly outside and pushed), the minimum distance achieved from the target is plotted in Fig.~(\ref{fig:acc}). }\\
\rev{Both }\del{All three} plots reflect the \del{stark} disparity in success rates. Figure~(\ref{fig:pick}) \del{shows}\rev{suggests } that with A and B, it was faster \emph{not} to operate the arm than to attempt picking using the method, while C and D enabled picking of many more blocks at comparable rates. For placing, B, C and D share a common trend of participants opting not to use the method slightly more often than to use it. The accuracy of placing\del{\footnote{Blocks which crossed the partition but did not reach the target (more often in A and B), are excluded here.}}, conditioned on success, shows no additional difference between methods\del{(Fig.~(\ref{fig:acc}))}. 
\del{This is unsurprising, as accuracy was given lowest priority, and the target regions were displayed \rev{clearly } (Fig.~(\ref{fig:sysarch}b)). }
\subsubsection{Questionnaire responses}
\begin{figure}[htpb]
\centering
\includegraphics[trim={2cm 0cm 0.45cm 1cm},clip,width=0.5\textwidth]{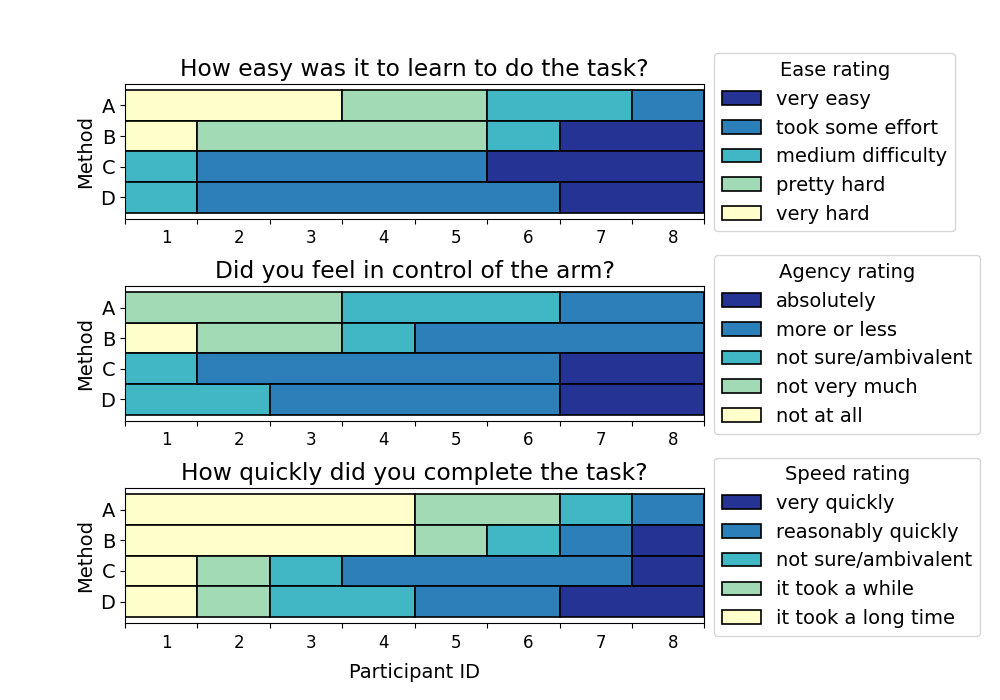}
\vspace{-2em}
\caption{Survey rating results after each method \rev{(Study~1)}.}
\label{fig:survey}
\end{figure}

Participants do not seem to have perceived any significant difference between C and D, but \del{unambiguously}preferred them over A and B~(Fig~(\ref{fig:survey}), Table~\ref{tb:ranks}). 
\begin{table}
\caption{Final survey responses ranking methods \rev{(Study~1)}. \rev{Identical answers are shaded in the same color.}}
\label{tb:ranks}
\begin{tabular}{|p{7pt}|p{55pt}|p{55pt}|p{55pt}|p{18pt}|}
\hline
P & Ease & Agency & Speed & Choice \\
\hline
1 & \cellcolor[HTML]{ccebc5} D $>$ C $>$ A $>$ B & D $>$ A $>$ C $>$ B & D $>$ A $>$ C $>$ B & D \\
2 & \cellcolor[HTML]{eff3ff} C $>$ D $>$ B $>$ A &	\cellcolor[HTML]{eff3ff} C $>$ D $>$ B $>$ A &	\cellcolor[HTML]{eff3ff} C $>$ D $>$ B $>$ A & C \\
3 & \cellcolor[HTML]{ccebc5} D $>$ C $>$ A $>$ B & \cellcolor[HTML]{ccebc5} D $>$ C $>$ A $>$ B &	\cellcolor[HTML]{ccebc5} D $>$ C $>$ A $>$ B &	D \\
4 & \cellcolor[HTML]{deebf7} D $>$ C $>$ B $>$ A & \cellcolor[HTML]{deebf7} D $>$ C $>$ B $>$ A & \cellcolor[HTML]{deebf7} D $>$ C $>$ B $>$ A &	D \\
5 & \cellcolor[HTML]{deebf7} D $>$ C $>$ B $>$ A &	\cellcolor[HTML]{deebf7} D $>$ C $>$ B $>$ A &	\cellcolor[HTML]{eff3ff} C $>$ D $>$ B $>$ A &	D \\
6 & \cellcolor[HTML]{edf8e9} D = C $>$ A $>$ B &	\cellcolor[HTML]{edf8e9} D = C $>$ A $>$ B &	\cellcolor[HTML]{edf8e9} D = C $>$ A $>$ B &	D \\ 
7 & \cellcolor[HTML]{eff3ff} C $>$ D $>$ B $>$ A &	C $>$ B $>$ D $>$ A &	\cellcolor[HTML]{deebf7} D $>$ C $>$ B $>$ A &	B \\ 
8 & \cellcolor[HTML]{ccebc5} D $>$ C $>$ A $>$ B &	\cellcolor[HTML]{ccebc5} D $>$ C $>$ A $>$ B &	\cellcolor[HTML]{ccebc5} D $>$ C $>$ A $>$ B &	D \\
\hline
\end{tabular}
\vspace{-1em}
\end{table}
\begin{figure*}[bhtp!]
\centering
\begin{subfigure}[t]{0.48\textwidth}
    \includegraphics[trim = {1cm 1cm 2cm 0}, clip, width=\textwidth]{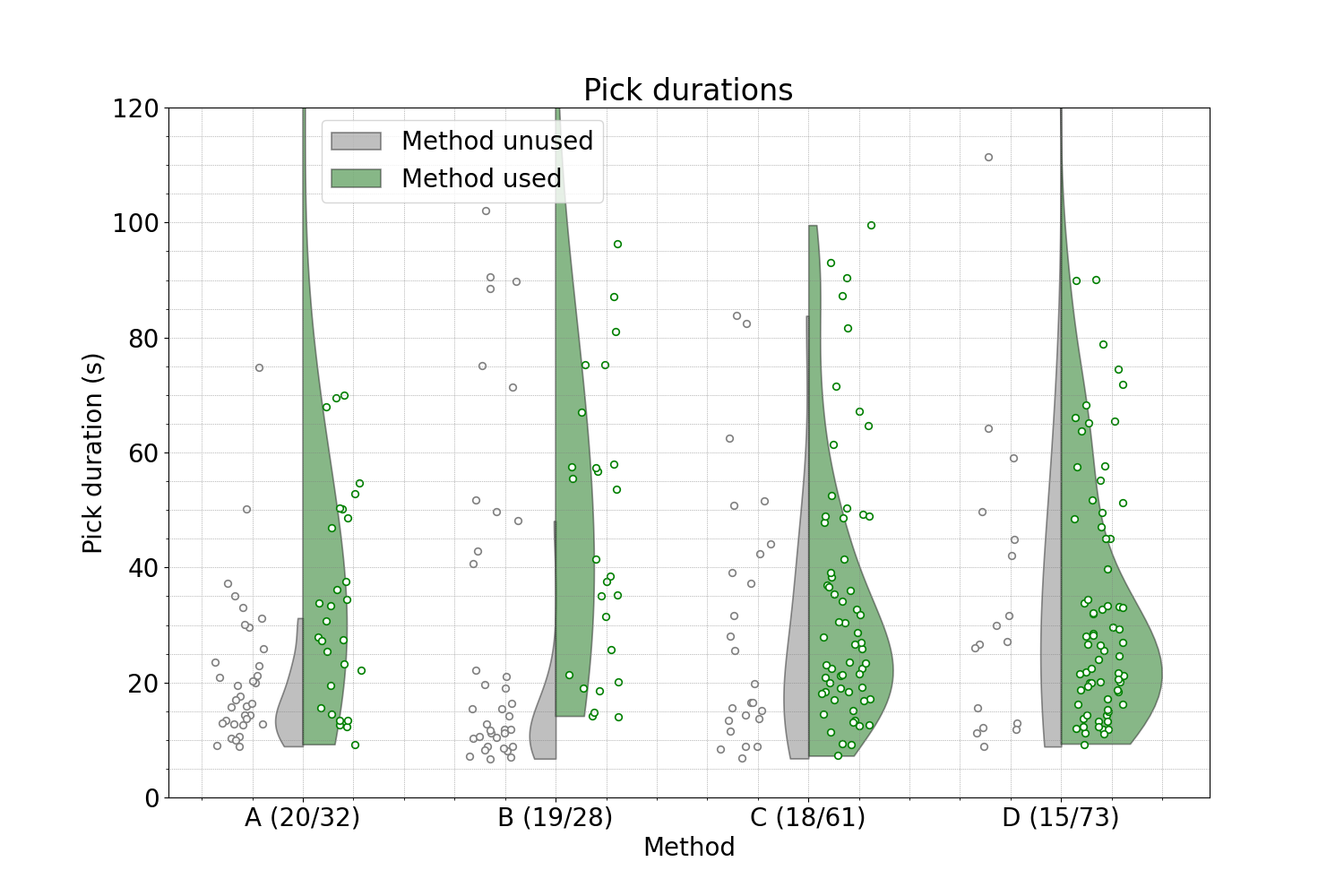}
    \caption{}
    \label{fig:pick}
\end{subfigure}
\begin{subfigure}[t]{0.48\textwidth}
    \includegraphics[trim = {1cm 1cm 2cm 0}, clip, width=\textwidth]{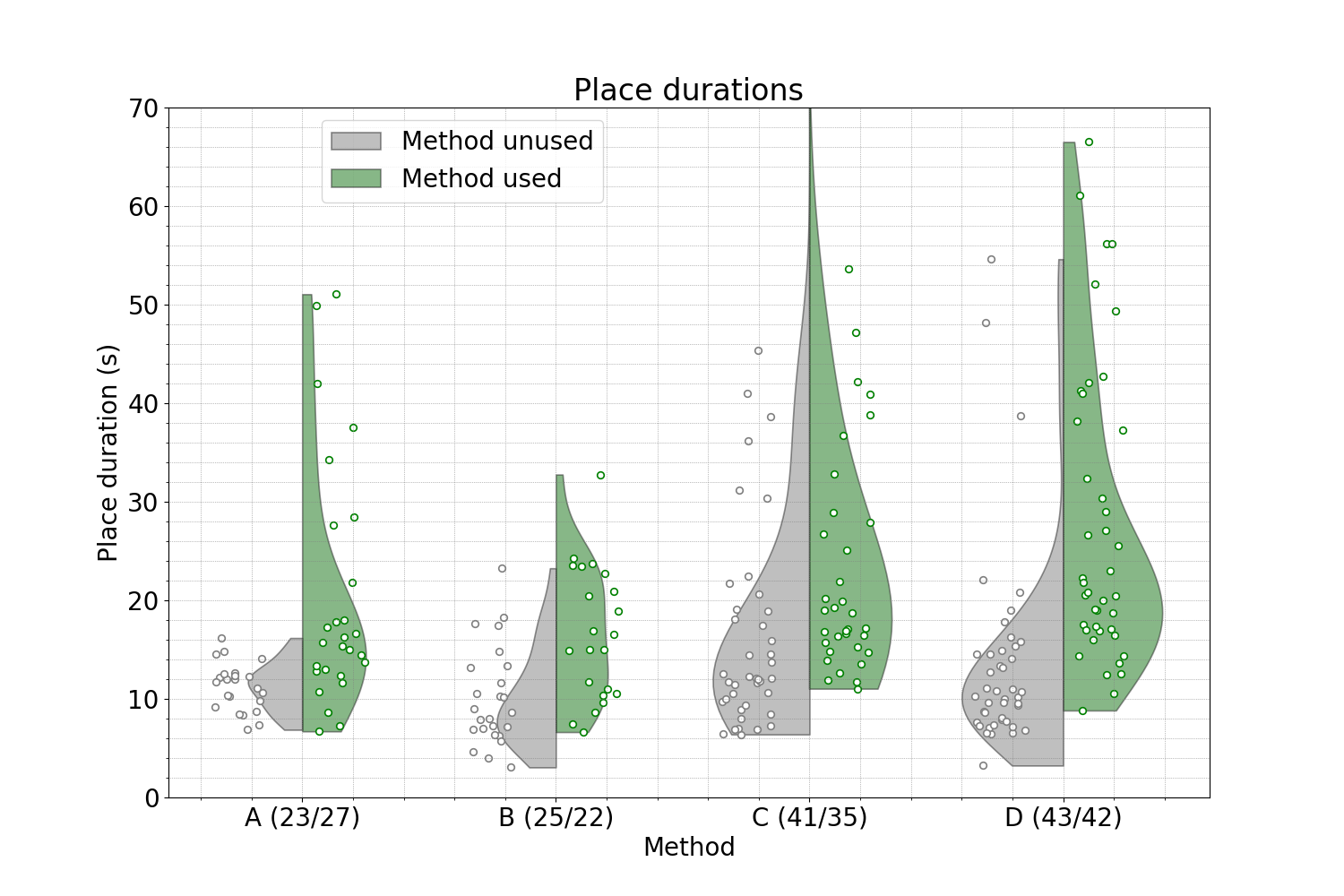}
    \caption{}
    \label{fig:place}
\end{subfigure}
\caption{Pick and place durations \rev{from Study~1}, excluding extreme outliers (lower is better; details in Section~\ref{sec:timings}). The numbers of actions performed without/with each provided method are noted along the ``Method" axis. }
\label{fig:pp}
\end{figure*}
\del{
\begin{figure}[hbtp]
\centering
\includegraphics[width=0.5\textwidth]{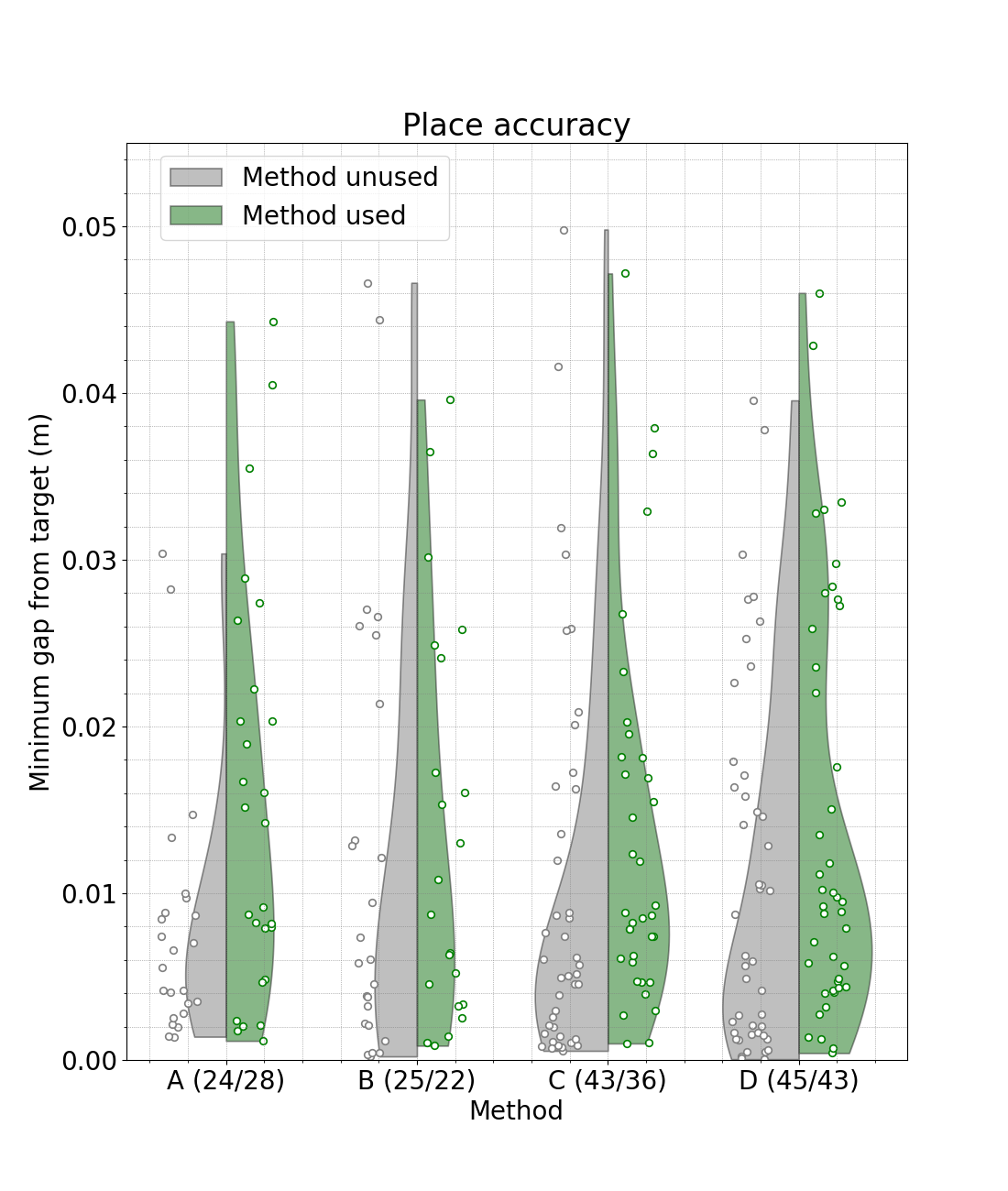}
\caption{Average of the minimum distance of block from target \rev{(Study~1)}. The numbers of place actions performed without/with each provided method are noted along the ``Method" axis.}
\label{fig:acc}
\end{figure}
}
\rev{
\subsection{Study 2}
Study~1 suggests that the two assisted methods might be conducive to performance when participants are first introduced to each method, but it does not capture the learning effect expected as participants gain more experience with each method. It also shows a clear effect of fatigue, albeit averaged across participants through counterbalancing. Therefore, Study~2 further examined performance with the two ``extremes"~(A and~D) in greater detail. \\Participants~1,~2, and~6 reported having some prior experience with EMG controllers. It was noted that~P3 and~P5 complained about EMG misclassification throughout method~D, with P3 taking about~$13$ minutes to retrain the classifier after trial 2. Data from these sessions were retained, as they show the effect of classification uncertainty on behavior and performance. Participant~4 also retrained the classifier after trial~2, but did not complain during subsequent trials\footnote{Note:~P4 later reported that until trial~8, he had not internalized that the wrist needed to be held extended for executing motion plans.}.  \\
During their respective second sessions, all three~(P3,~P4,~P5) were able to achieve highly satisfactory classification (better than any previous instance) within~$1$-$3$ attempts. Results show a split between participants whose performance seemed to benefit from method~D, and those who performed nearly the same with both methods. For clarity, those who show similar trends~(P2,~P4,~P6) are referred to as Group~1 below. The members of Group~2~(P1,~P3,~P5) are addressed individually. 
\subsubsection{Success rates}
The total numbers of successful transfers per trial, progressing across trials, are shown in Fig.~\ref{fig:success2}. Group~1 participants showed higher success rates with~D than with~A. 
\begin{figure*}[bhtp!]
\centering
\begin{subfigure}[t]{\textwidth}
    \includegraphics[width=\textwidth]{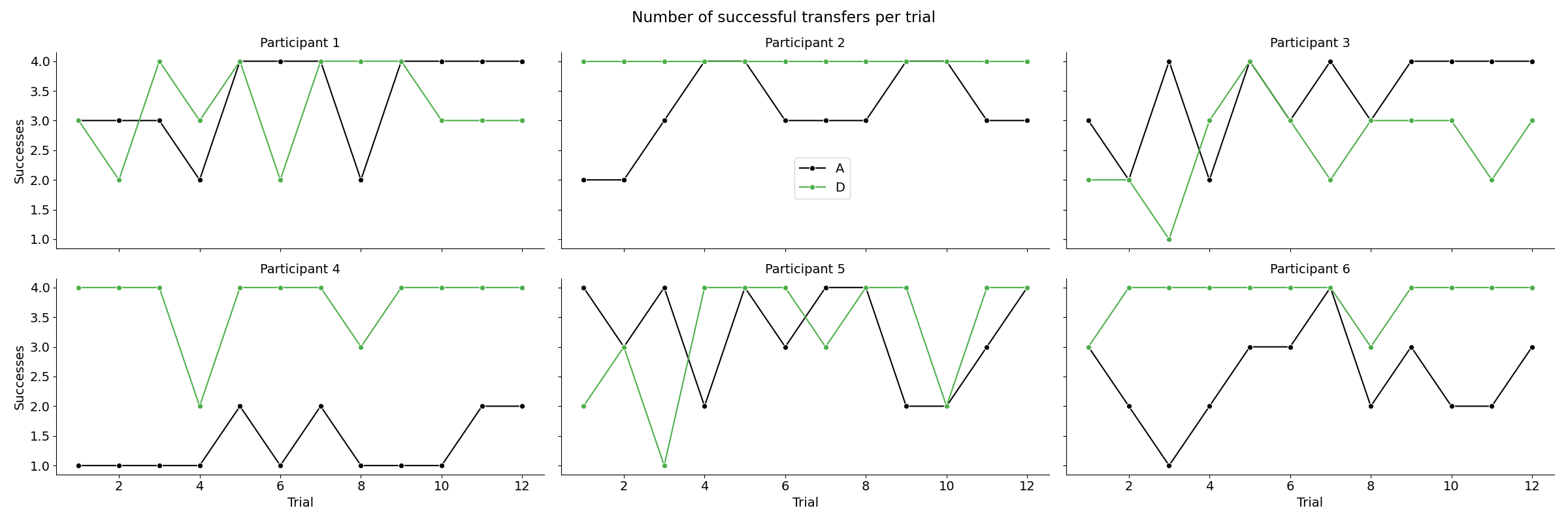}
\end{subfigure}
\caption{\rev{Success rates across trials \rev{(Study~2)}; participants~2,~4, and~6 did consistently better with~method D, while others' performance fluctuated.} }
\label{fig:success2}
\end{figure*}
\subsubsection{Transfer timings and accuracy}
As participants were explicitly asked to use the provided methods as much as possible, we did not make similar annotations as in Fig.~\ref{fig:pick}. Group~1 seem to have performed pick actions somewhat faster and more consistently with Method~D (Fig.~\ref{fig:pick2}), as well as placing blocks with somewhat higher accuracy and greater consistency, including that fewer blocks were disturbed after having entered the target region (Fig.~\ref{fig:gap2}). Group~2 seem to perform about the same with either method.
\begin{figure*}[bhtp!]
\centering
\begin{subfigure}[t]{0.49\textwidth}
    \includegraphics[width=\textwidth]{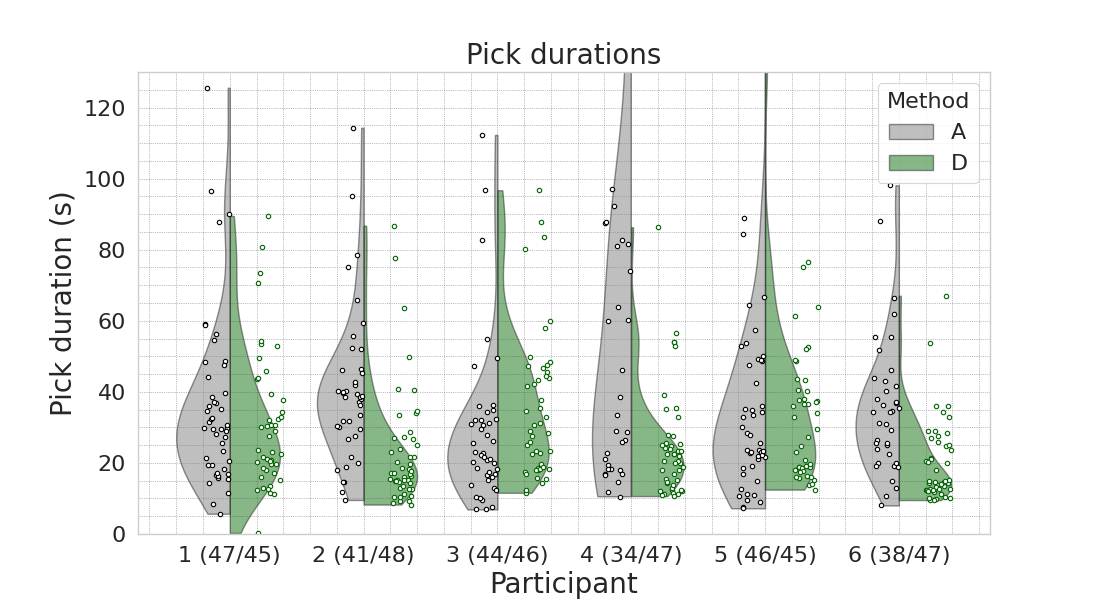}
    \caption{}
    \label{fig:pick2}
    \vspace{-7em}
\end{subfigure}
\begin{subfigure}[t]{0.49\textwidth}
    \includegraphics[width=\textwidth]{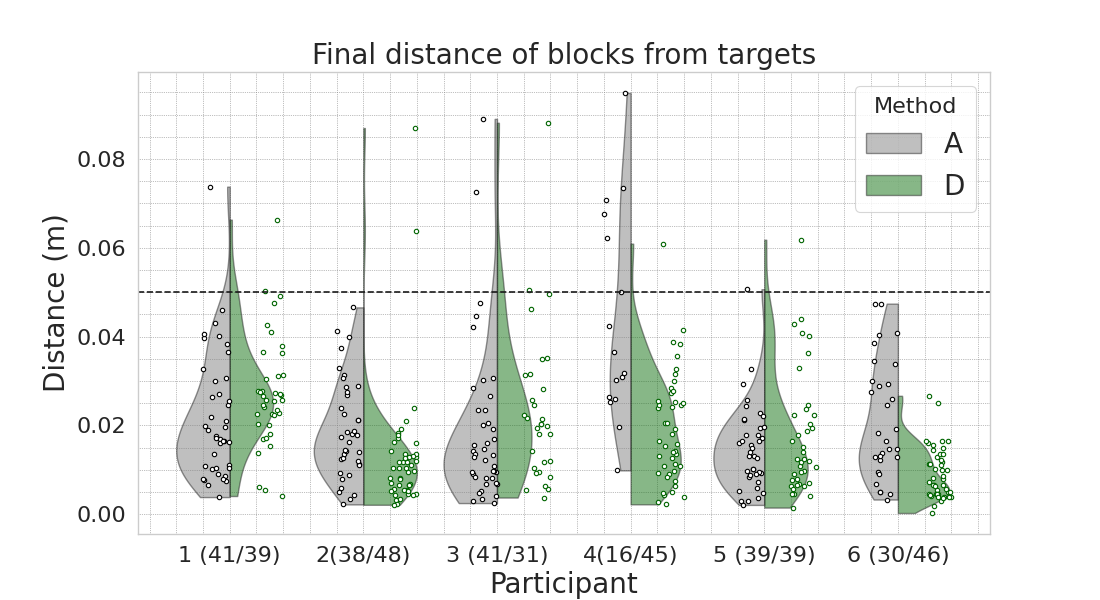}
    \caption{}
    \label{fig:gap2}
    \vspace{-7em}
\end{subfigure}
\begin{subfigure}[t]{0.49\textwidth}
    \includegraphics[width=\textwidth]{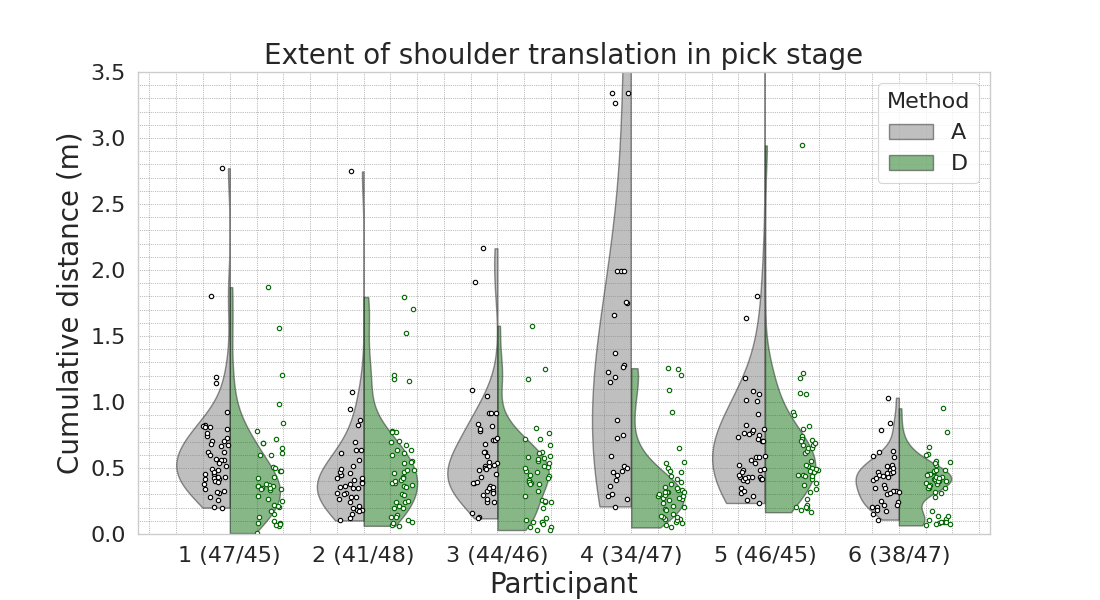}
    \caption{}
    \label{fig:lins}
\end{subfigure}
\begin{subfigure}[t]{0.49\textwidth}
    \includegraphics[width=\textwidth]{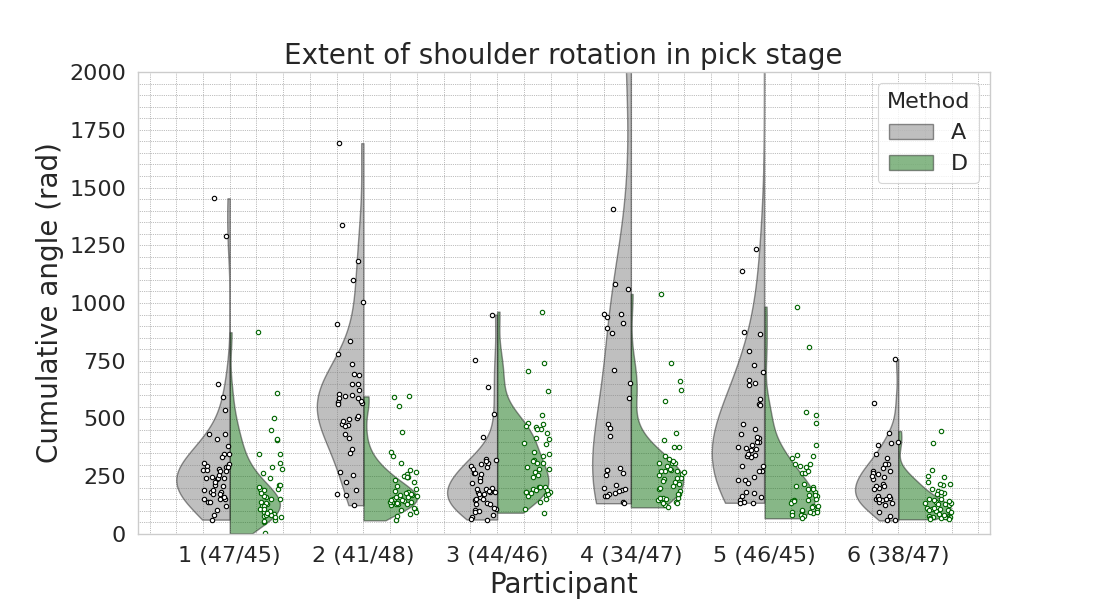}
    \caption{}
    \label{fig:angs}
\end{subfigure}
\caption{\rev{Detailed plots showing speed and compensatory motion during pick and place stages \rev{(Study~2)}.} }
\end{figure*}
\subsubsection{Compensatory movement}
The cumulative motion of the shoulder coordinate frame was calculated for each successful pick and place. Rotation was quantified by computing the angle of rotation (about the appropriate axis) in each time step at~$10$~Hz and summing up these angles; therefore, its absolute value, which depends upon the time step, is not meaningful, but the relative values can be compared. In Fig.~\ref{fig:angs}, Group~1 appear to have used the rotation of the shoulder to a greater extent when using method~A. Perhaps because translation is easier to notice and avoid, the translation data do not show a similar trend (Fig.~{\ref{fig:lins}}). The slightly greater rotation by~P5 and translation by~P1 in~A may partially explain their comparable performance between~A and~D along the above metrics. 
\subsubsection{Gaze}
Between the two methods, participants spent a similar fraction of the total transfer time looking at the arm (including mode display). Within the parts of the arm, it is interesting to note that for all participants, more or less consistently across trials, visual attention was more often directed towards the hand in~D when compared to~A. Additionally, it was observed that (likely because~D includes stages when gaze is actively engaged) the fraction of time spent looking at place targets is greater with~D.  
\subsubsection{Questionnaire responses}
Weighted scores from Pros-TLX 
show a mix of results. Conscious processing and visual attention required were perceived to be higher in~A by most participants, with uncertainty being higher in~D. Overall, all participants except~P3 perceived higher workload in~A. 
}
\section{Discussion}
\rev{
Results from both studies show that, at least for a section of the (small) sample studied, gaze-based assistance can provide a notable improvement in performance. More importantly, the proposed platform not only also illuminates the large variation among participants that is typical of user studies with assistive technology, but allows us to examine factors which led to improvement or degradation in performance -- whether for an individual participant, individual trial, or individual object interaction -- and to iteratively design better control strategies.\\
The ability to track compensatory motions and isolate actions relying upon them is a valuable tool in design. For example, Fig.~\ref{fig:place} made it clear that~C and~D were not helpful to participants in the place stage. Anecdotally, this could be explained by the specific implementation of the placing method. Multiple participants reported difficulty in making the gaze-tracking marker appear on the target side, during the first placing action of trials. If they either succeeded, or decided to use body movement instead, then subsequent transfers were completed quickly. By using the testbed to highlight this flaw, we were able to correct it before Study~2, by preemptively displaying the marker. We were also able to identify that the primary bottleneck for successful transfers in this task was not placing, but reaching to pick, which requires more precise control (Fig.~(\ref{fig:pp})), explaining why the methods which improve picking might have led to higher success rates. }\\
It is for a similar reason that the differences between C and D are not obvious from this study; reaching to pick, and not selecting the object to be picked, was the major challenge, which both methods alleviated similarly. Future work could investigate cluttered scenes, in which mere proximity of the center of gaze to different objects might be insufficient. \\
The nature of the \gls{BBT} allowed P1 in Study~1 to achieve very fast transfers through skilled movement of the upper body as a whole. While all participants were informed that they were free to move thus, and in fact, would need to do so in order to reach all blocks, only one was able to use body movement to such an extent and with such skill. Had all adopted this strategy, no results of interest might have been obtained. \rev{This motivates future studies based on a more spatially complex task (such as the PHAM~\cite{pham, baskarHoloPHAMAugmentedReality2017} or the pasta box task~\cite{valeviciusCharacterizationNormativeHand2018}). \\
All our participants had had at least occasional prior exposure to video games. It was found that users na\"ive to EMG who also have \emph{no} experience of video games struggle to use hand gestures as independent controls. This is a limitation of the studies in their current form; it is not clear whether a slow-paced ``zeroth" session might address it. For participants like P3-P5 in Study~2, who were able to pass the set criteria for classification but still struggled with control during their first session, there is no doubt that a preliminary session before the first day (as well as a more stringent classification test, such as~$75\%$ accuracy in a longer trial~\cite{williams2024}) would allow for a fairer comparison between methods. \\
As participants were aware of the nature of each method, it is difficult to judge whether the minor improvement in Study~1 with D relative to C in success rates may be attributed to participant bias. }However, as D occurs before and after C the same number of times in Study~1, learning and fatigue can both be ruled out as explanations. The effect of modeling sequential tasks merits further investigation with a larger dataset. \\
Six of eight participants selected D as the method they would most prefer if attempting an additional trial. P2 selected C, which is explained by the deterioration of \gls{emg} acquisition in the first iteration of D (Table~\ref{tb:seq}). 
P7 selected B, explaining that, even though it was more difficult than C/D, it made him feel the most satisfied on success, because he felt in control. This is especially interesting in light of the fact that during trials with B, P7 had declared a general weakness in spatial reasoning, which was most challenged by end-effector control. Understandably, P7 alone ranked B above D on the question of agency. P1 ranked A as second in agency and perceived speed, which may be explained by the technical issue in the first iteration of C. Aside from these, all participants in Study~1 ranked C and D as the top two in all three metrics: ease, agency, and perceived speed. \rev{The design choice to move the arm along the planned trajectory \emph{only as long as \gls{emg} input was actively provided} seems to have been sufficient for most participants' perceived sense of control. While these studies do not constitute a thorough analysis of agency or embodiment, which are crucial for prosthesis users, ProACT might serve as a tool for such studies in the future. \\
In the comparison between A and B, exactly half of the participants preferred A over B and vice-versa. At first glance, this may appear to be a natural variation in preferences. However, examining Tables~(\ref{tb:seq},~\ref{tb:ranks}) in conjunction reveals that each participant preferred whichever of these two methods they had performed first; very likely a fatigue effect. It is encouraging that in the light of such an effect, the improvements offered by C and D (which were encountered in alternation, as shown in Table~\ref{tb:seq}) overwhelmed fatigue to keep them at the top of nearly all rankings. \\
Study~2 indicates the need for further improvements, e.g., feedback about the status of the planner, and a double confirmation for opening the hand to reduce the risk of dropping. Those who faced uncertainty in EMG control also remarked that biofeedback would be helpful in distinguishing it from the uncertainty of planning success. \\
It is of interest to study
the effect of methods which use gaze, whether actively or
passively, upon gaze behavior during the task. For example,
P1 in Study 2, looked at the upper arm more often in A than
in D, the additional visual attention, by his own description,
aiding in control. All participants looking at the hand often during~D also suggests that they were more unsure of being able to predict its behavior, due to the autonomous motion planning. Following~\cite{parr2017}, gaze can be developed into a useful tool for evaluating the level of trust/certainty when using methods in the platform. \\
ProACT can also be used to further improve conventional methods like joint-based control, e.g., by studying which joints are used more heavily in tasks, and informing decisions about which subset of modes to cycle through, as is already done in commercial prostheses, or intelligently select and automate certain joints at certain times. Indeed, some users might prefer such a form of partial autonomy, as compared to task-space motion planning. It is through performing rigorous studies with large and generalizable samples that these questions might be answered; it is our hope that ProACT will enable such investigations. \\  
}

\label{sec:discussions}
\del{
The large variation among participants seen in the above results is a typical feature of user studies with assistive technology, which is encouraging to see in this new platform. The excellent performance of Participant 1 (P1) across all methods demonstrates that the task was designed to be possible to complete within the allotted time; at the same time, the remaining participants exhibit a wide range of skill: P8 had a relatively poor success rate, close to~$50\%$, across methods, while P2 and P5 showed a large improvement in performance between methods A and D (Fig.~(\ref{fig:fatedetail})). \\
The testbed highlights the strengths of the new methods -- pick times are reduced and made more consistent -- which allowed participants to achieve higher success rates, while place times were not improved, and in some cases, actually increased. Anecdotally, this can be explained by the specific implementation of the placing method in C and D. Multiple participants reported difficulty in making the gaze-tracking marker appear on the target side, during the first placing action of trials. If they either succeeded, or decided to use body movement instead, then subsequent transfers were completed quickly. By using the testbed to highlight this failure of the methods being prototyped, we are able to identify an area of future work. We are also able to identify that the primary bottleneck for successful transfers in this task is not placing, but reaching to pick, which requires more precise control (Fig.~(\ref{fig:pp})), explaining why the methods which improve picking lead to unambiguously higher success rates. \\
It is for a similar reason that the differences between C and D are not obvious from this study; reaching to pick, and not selecting the object to be picked, was the major challenge, which both methods alleviated similarly. In future work, we shall investigate cluttered scenes, in which mere proximity of the center of gaze to different objects might be insufficient. \\
The nature of the \gls{BBT} allowed P1 to achieve very fast transfers through skilled movement of the upper body as a whole. While all participants were informed that they were free to move thus, and in fact, would need to do so in order to reach all blocks, only one was able to use body movement to such an extent and with such skill. Had all participants adopted this strategy, no significant results might have been obtained. This motivates future studies based on a more spatially complex task, where objects would not lie on the same plane, and the intended direction of their displacement would not be along a single axis. \\
The small sample size of this study and the disparate success rates precluded statistical analyses of performance metrics and survey responses, but results suggest that, given the notable random effect of participant skill, mixed-effects models would be most appropriate for larger datasets in the future, with Friedman tests for the pseudo-quantitative components.\\
No deception was used in this study; participants were possibly biased in favor of C and D, perhaps slightly favoring D. It is difficult to judge whether the minor improvement with D relative to C in success rates may be attributed to any assumed superiority. However, as D occurs before and after C the same number of times, learning and fatigue can both be ruled out as explanations. The effect of modeling sequential tasks merits further investigation with a richer dataset. \\
Six of eight participants selected D as the method they would most prefer if attempting an additional trial. P2 selected C, which is explained by the deterioration of \gls{emg} acquisition in the first iteration of D (Table~\ref{tb:seq}). 
P7 selected B, explaining that, even though it was more difficult than C/D, it made him feel the most satisfied on success, because he felt in control. This is especially interesting in light of the fact that during trial with B, P7 had declared a general weakness in spatial reasoning, which was most challenged by end-effector control. Understandably, P7 alone ranked B above D on the question of agency. P1 ranked A as second in agency and perceived speed, which may be explained by the technical issue in the first iteration of C. Aside from these, all participants ranked C and D as the top two in all three metrics: ease, agency, and perceived speed. The choice to move the arm along the planned trajectory \emph{only as long as \gls{emg} input was actively provided} seems to have been sufficient to provide a sense of agency to most participants. \\
More interestingly, in the comparison between A and B, exactly half of the participants preferred A over B and vice-versa. At first glance, this may appear to show some natural variation in preferences. However, examining Tables~(\ref{tb:seq},~\ref{tb:ranks}) in conjunction reveals that each participant preferred whichever of these two methods they had performed first; very likely a fatigue effect. It is encouraging that in the light of such a strong effect, the improvements offered by C and D (which were encountered in alternation, as shown in Table~\ref{tb:seq}) overwhelmed the effect of fatigue to keep them at the top of nearly all rankings. \\}
\section{Conclusion}
\label{sec:conclusion}
\rev{T}he present work demonstrates the following:
\begin{enumerate}
    \item Gaze-based intent estimation and low-level autonomy \rev{have the potential to } improve the performance and comfort of users of high-\gls{dof} prosthetic arms in complex manipulation tasks\rev{, based on preliminary data from non-amputees}. 
    \item The \gls{this} testbed is useful in prototyping evaluation tasks as well as control/feedback methods, identifying specific features of methods (and phases of tasks) which contribute positively or negatively to performance, and studying variation within and across participants. 
\end{enumerate}
Due to the open-source nature of much of the latest robotics infrastructure, as well as \gls{this} itself, it is easy to incorporate more sophisticated methods of intent prediction, and test them in the context of more challenging tasks, a different level of limb loss, or other wearable robotic manipulators. \rev{Future work with greater powered samples, as well as with amputee populations, will provide some of the insights necessary for the realization of intelligent whole-arm prostheses.}
\section{References}
\errorcontextlines=99
\bibliography{main}{}
\bibliographystyle{IEEEtran}
\end{document}